\documentclass[conference]{IEEEtran}
\IEEEoverridecommandlockouts
\usepackage{cite}
\usepackage[normalem]{ulem}
\usepackage[numbers]{natbib}
\usepackage{graphicx}
\usepackage{amsmath,amssymb,amsfonts}
\usepackage{algorithm}
\usepackage{algorithmic}
\usepackage{graphicx}
\usepackage{textcomp}
\usepackage{xcolor}
\usepackage{booktabs}
\usepackage{url}
\usepackage{hyperref}
\def\BibTeX{{\rm B\kern-.05em{\sc i\kern-.025em b}\kern-.08em
    T\kern-.1667em\lower.7ex\hbox{E}\kern-.125emX}}
\begin{document}

\title{A New Perspective on Time Series Anomaly Detection: Faster Patch-based Broad Learning System}

\author{\IEEEauthorblockN{Pengyu Li\IEEEauthorrefmark{2}}
\IEEEauthorblockA{
\textit{South China University of Technolegy} \\
Guangzhou, China \\
202421044678@mail.scut.edu.cn}
\and
\IEEEauthorblockN{Zhijie Zhong\IEEEauthorrefmark{2}}
\IEEEauthorblockA{
\textit{South China University of Technology}\\
Guangzhou, China \\
csemor@mail.scut.edu.cn}
\and
\IEEEauthorblockN{Tong Zhang}
\IEEEauthorblockA{
\textit{South China University of Technology}\\
Guangzhou, China \\
tony@scut.edu.cn}
\and
\IEEEauthorblockN{Zhiwen Yu}
\IEEEauthorblockA{
\textit{South China University of Technology}\\
Guangzhou, China \\
zhwyu@scut.edu.cn}
\and
\IEEEauthorblockN{C.L.Philip Chen}
\IEEEauthorblockA{
\textit{South China University of Technology}\\
Guangzhou, China \\
philip.chen@ieee.org}
\and
\IEEEauthorblockN{Kaixiang Yang\IEEEauthorrefmark{1}}
\IEEEauthorblockA{
\textit{South China University of Technology}\\
Guangzhou, China \\
yangkx@scut.edu.cn}
\IEEEcompsocitemizethanks{
\IEEEcompsocthanksitem \IEEEauthorrefmark{2}Equal contribution.
\IEEEcompsocthanksitem \IEEEauthorrefmark{1}Corresponding author: Kaixiang Yang.
}
}

\maketitle

\begin{abstract}
Time series anomaly detection (TSAD) has been a research hotspot in both academia and industry in recent years. Deep learning methods have become the mainstream research direction due to their excellent performance. However, new viewpoints have emerged in recent TSAD research. Deep learning is not required for TSAD due to limitations such as slow deep learning speed. The Broad Learning System (BLS) is a shallow network framework that benefits from its ease of optimization and speed.
It has been shown to outperform machine learning approaches while remaining competitive with deep learning. Based on the current situation of TSAD, we propose the \textbf{C}ontrastive \textbf{Patch}-based \textbf{B}road \textbf{L}earning \textbf{S}ystem (\textbf{CPatchBLS}). This is a new exploration of patching technique and BLS, providing a new perspective for TSAD. We construct Dual-PatchBLS as a base through patching and Simple Kernel Perturbation (SKP) and utilize contrastive learning to capture the differences between normal and abnormal data under different representations. To compensate for the temporal semantic loss caused by various patching, we propose CPatchBLS with model-level integration, which takes advantage of BLS's fast feature to build model-level integration and improve model detection. Using five real-world series anomaly detection datasets, we confirmed the method's efficacy, outperforming previous deep learning and machine learning methods while retaining a high level of computing efficiency.

\end{abstract}

\begin{IEEEkeywords}
Ensemble learning, Broad Learning System (BLS), data mining, and time series anomaly detection.
\end{IEEEkeywords}

\section{Introduction}

Time series anomaly detection (TSAD) is a key task in time series analysis, playing a crucial role in various fields such as finance, personal health, and industry. \cite{BLS_AdaMemBLS2024}
The economic value inherent in various fields has led to the widespread development of TSAD in recent years. Researchers initially explored the use of traditional machine learning methods that are intuitive and convenient, such as LOF and Isolation Forest. Subsequently, deep learning models are delved due to strong performance and greater generalization capabilities, such as TCN and Transformer \cite{CL_AnomalyTransformer2021,intro_jeong2023anomalybert,CL_DCdetector2023,simads,simad}. More recently, the deep learning route based on MLP \cite{Patch_PatchAD2024,Patch_PatchMixer2023} has also gained attention. More effective and powerful deep learning detection models are explored and developed. 
\textbf{However, is the deep learning architecture a necessity in all practical application scenarios for implementing TSAD? Are there other perspectives to implementing TSAD?}

In the realm of time series prediction, \citet{intro_zeng2023transformers} demonstrated that the multi-head attention mechanism in Transformers does not adeptly adapt to temporal data. Conversely, they proposed a simple model, DLinear, that exhibits greater robustness and is less susceptible to noise interference. Currently, there is still skepticism within the community regarding whether the accomplishments of Transformer architectures in NLP can be readily extended to diverse time series tasks.
Moreover, \citet{intro_sarfrazposition} have raised concerns regarding deep learning, emphasizing that while deep learning has shown promising performance in the domain of TSAD, existing studies have not provided sufficient evidence regarding the indispensability of deep learning. Therefore, the necessity of deep learning also requires further investigation.

Currently, the TSAD community faces a dilemma where traditional machine learning methods struggle to capture temporal information and anomaly semantics, while deep learning methods are slow in speed, and their necessity is being questioned. This is particularly critical in industrial scenarios, where there is a pressing need for swift algorithms.
\textbf{Is there a more balanced option that preserves the robustness and speed advantages of machine learning while also demonstrating the powerful representational capacity akin to deep learning?}

To address the aforementioned issue, we grounded in the Broad Learning System (BLS) and proposed a potential alternative solution called \textbf{C}ontrastive \textbf{Patch}-based \textbf{B}road \textbf{L}earning \textbf{S}ystem (\textbf{CPatchBLS}), which is designed for unsupervised TSAD. CPatchBLS is a BLS architecture that deviates from deep learning frameworks by updating parameters via pseudo-inverse computation, eliminating away with the need for backpropagation. CPatchBLS achieves accurate time series anomaly detection with accelerated training and testing speeds while also offering a simpler model structure.

In specific terms, the contributions of this paper can be delineated into the following three points:
\begin{enumerate}
    \item To model temporal data, we introduce the concept of patching into BLS for the first time, proposing PatchBLS to enhance the capability of BLS in extracting temporal semantic information.
    \item The dissimilarity is utilized between normal and anomalous views for TSAD, and Simple Kernel Perturbation (SKP)-PatchBLS is incorporated to construct Dual-PatchBLS for anomaly identification.
    \item Harnessing the rapid advantages of BLS and integratin multi-scalele features for robust representatio, CPatchBLS maintains the benefits of both machine learning and deep learning, surpassing previous deep learning methods across multiple datasets.
\end{enumerate}

\begin{figure*}[ht]
\centering
\includegraphics[width=1\linewidth]{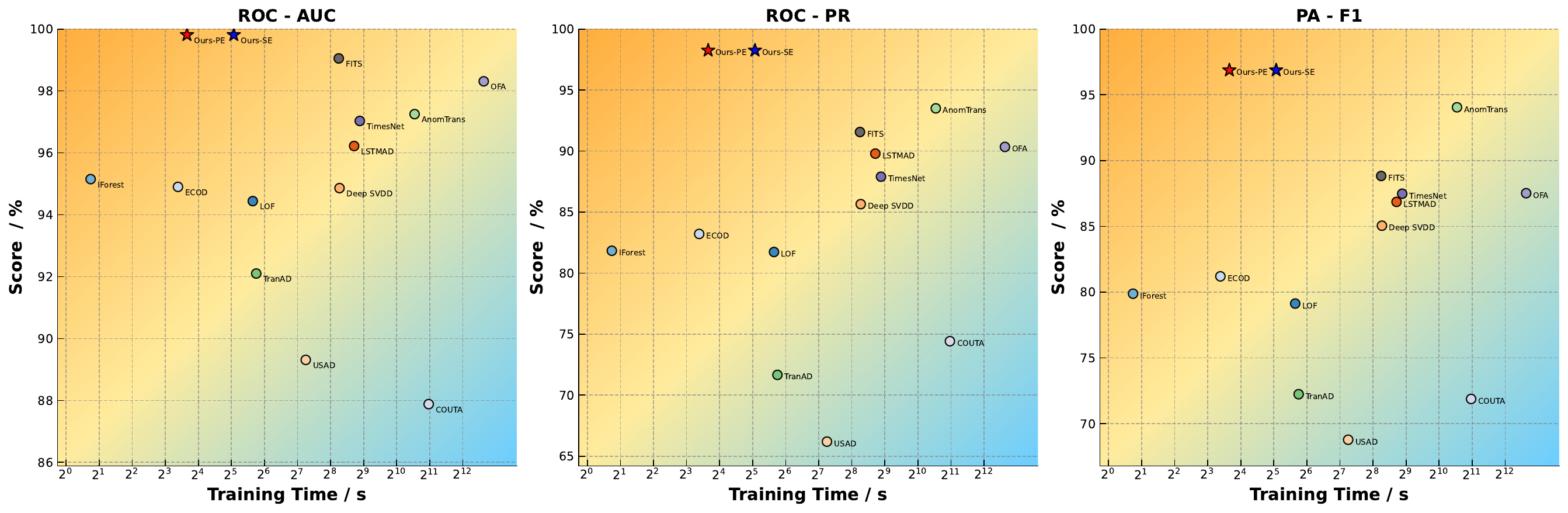}
\caption{By analyzing key metrics like ROC-AUC, PR-AUC, and PA-F1 for twelve state-of-the-art models and our methods on five datasets, the figure shows an association between training time and effectiveness. Models closer to the top-left in the heatmap (indicated by darker red shades) achieve a superior trade-off between time efficiency and performance, whereas those near the bottom-right (darker blue shades) indicate suboptimal performance or higher time consumption. Our methods, represented by red pentagonal stars (\textbf{Ours-PE}, parallel execution) and blue pentagonal stars (\textbf{Ours-SE}, serial execution), are positioned closest to the top-left red corner among the fourteen approaches. This demonstrates clear advantages of ours in both time efficiency and performance, highlighting the competitiveness against advanced baselines.}
\label{1}
\end{figure*}

\section{Related Work}

\subsection{Traditional TSAD} 

Traditional TSAD methods are improved from classical machine learning methods by extending them to time series scenarios.
Autoregressive Integrated Moving Average (ARIMA) is a forecasting method using historical data to predict future patterns, with the assumption that higher prediction errors indicate anomalies. 
One-Class Support Vector Machine (OCSVM) \cite{ml_OCSVM2001} employs a classification-based TSAD approach, utilizing support vector machine principles. Isolation Forest (IForest) \cite{ml_IsolationForest2008}
constructs tree structures and computes the average path length across all trees for each sample, serving the path length as the anomaly score to complete anomaly detection.
Local Outlier Factor (LOF) \cite{ml_lof2000} assesses the ratio of local and neighboring densities, utilizing them for detecting anomalies. K-Nearest Neighbors (KNN) \cite{ml_KNN2000} assesses the degree of outlierness by measuring the proximity to the k-th nearest neighbor. 

Traditional classical methods like prediction, classification, tree partitioning, density, \textit{etc.}, can effectively detect anomalies in time series. Nevertheless, as industrial application automation rises and control systems become more complex, traditional TSAD methods encounter challenges. The scarcity of fault patterns results in label shortages, and the increasing complexity of time series data leads to the curse of dimensionality, causing performance degradation of classical algorithms \cite{ml_survey2021}.

\subsection{Broad Learning System TSAD}

The Broad Learning System (BLS) is an advanced model founded on Random Vector Functional Link Neural Networks (RVFLNN) \cite{BLS_2017}. BLS employs randomly initialized feature nodes and enhancement nodes for feature extraction. It learns the data distribution by solely updating the parameters of the output layer through pseudo-inverse calculation.
Unlike traditional deep learning, BLS addresses issues such as parameter tuning difficulties, slow training speeds, and the complexity of finding the optimal solution space.

AdaMemBLS \cite{BLS_AdaMemBLS2024} combines the fast inference capabilities of BLS with memory prototypes to effectively distinguish between normal and anomalous data. AdaMemBLS enhances the ability to learn time series features through various data augmentation techniques and incremental learning algorithms. Additionally, it employs diverse ensemble methods and discriminative anomaly scores to improve detection performance. And a theoretical analysis of the algorithm's time complexity and convergence demonstrates the feasibility of AdaMemBLS.

Despite the introduction of AdaMemBLS, its efficacy is constrained by the incapacity of the proposed temporal slices to adequately model temporal data. Our research suggests that the untapped potential of BLS in TSAD tasks warrants further exploration. In our study, we introduce PatchBLS and Contrastive BLS as extensions to the foundational BLS framework, aiming to augment the performance of BLS in TSAD and facilitate rapid anomaly detection within the BLS paradigm.

\subsection{Deep Learning TSAD}

\subsubsection{Patch Technique}

PatchTST \cite{Patch_PatchTST2022} utilizes a patch-based approach to model temporal sequences, emphasizing sequence-based modeling for time series data. It introduces patching and channel-independence mechanisms to capture sequential information, resulting in reduced computational complexity and improved performance over previous Transformer-based models. PatchTST demonstrates strong capabilities in multi-variable time series forecasting, with the effectiveness of the patching mechanism validated through ablation studies. 

In time series forecasting, PatchMixer \cite{Patch_PatchMixer2023} introduces a novel convolutional structure that replaces the computationally intensive self-attention in Transformer architectures with a single-scale depthwise separable convolutional block. This design optimizes patch representation and effectively captures temporal semantic information within and across patches. As a result, PatchMixer enhances the extraction of comprehensive semantic features, showcasing superior performance. This shows the potential of the patch technique for integrating with base models. It also demonstrates its adaptability to various network structures in time series analysis tasks.

In time series representation learning, PITS \cite{Patch_PITS2023} focuses on patch independence by ignoring inter-patch correlations. Instead of predicting masked patches based on unmasked data, it reconstructs unmasked patches independently. Utilizing contrastive learning, PITS generates complementary augmented views of original samples. By integrating reconstruction with hierarchical contrastive loss, it establishes an effective self-supervised learning framework, achieving strong performance across various time series tasks, including forecasting and classification.

Previous deep learning time series analysis research has gained a deeper understanding of the patching technique and believes that it also has a function similar to convolution and can extract basic features of time series. This characteristic can precisely compensate for the deficiency of BLS in time series feature extraction because BLS is designed for classification problems. As far as we know, we are the first to propose an end-to-end training model that combines patching and BLS.

\subsubsection{Contrastive Learning}

Anomaly Transformer \cite{CL_AnomalyTransformer2021} effectively detects anomalies in multi-variable time series by comparing global temporal attention patterns using series-association and prior-association with a Gaussian kernel, which signifies adjacent field attention patterns.
DCdetector \cite{CL_DCdetector2023} introduces a dual-branch attention-based contrastive learning structure, assuming that all normal data in the time series exhibit a shared latent temporal pattern. Researchers posit that normal data are easily represented consistently from various viewpoints, unlike anomalies. This approach aims to identify and explore anomalies within the time series data by comparing various representations within and across patches.

In multi-variable TSAD, \citet{Patch_PatchAD2024} introduce PatchAD, a lightweight architecture that employs contrastive learning and Patch-based MLP-Mixer. Building on PatchTST concepts, PatchAD extracts temporal semantic features using patches as core units. It utilizes distinct sub-models for different patches and integrates Patch Mixer encoder to capture complex relationships within and between patches and channels. A shared MixRep Mixer represents the feature space uniformly, while a dual project constraint mechanism helps prevent model collapse.

However, findings from \citet{Patch_PatchTST2022} and the additional constraint in PatchAD indicate that deep learning models with channel-mixing mechanisms often encounter problems such as poor generalization, convergence issues, overfitting, and pattern collapse, necessitating additional mechanisms that may introduce redundancy. In contrast, the proposed CPatchBLS incorporates a patching module for individual time series variables, effectively addressing these challenges through a multi-scale patch mechanism at the model integration level. The shallow network structure of BLS offers a faster, lightweight, and stable solution for TSAD, distinguishing it from previous models.

\begin{figure*}[ht]
\centering
\includegraphics[width=1\linewidth]{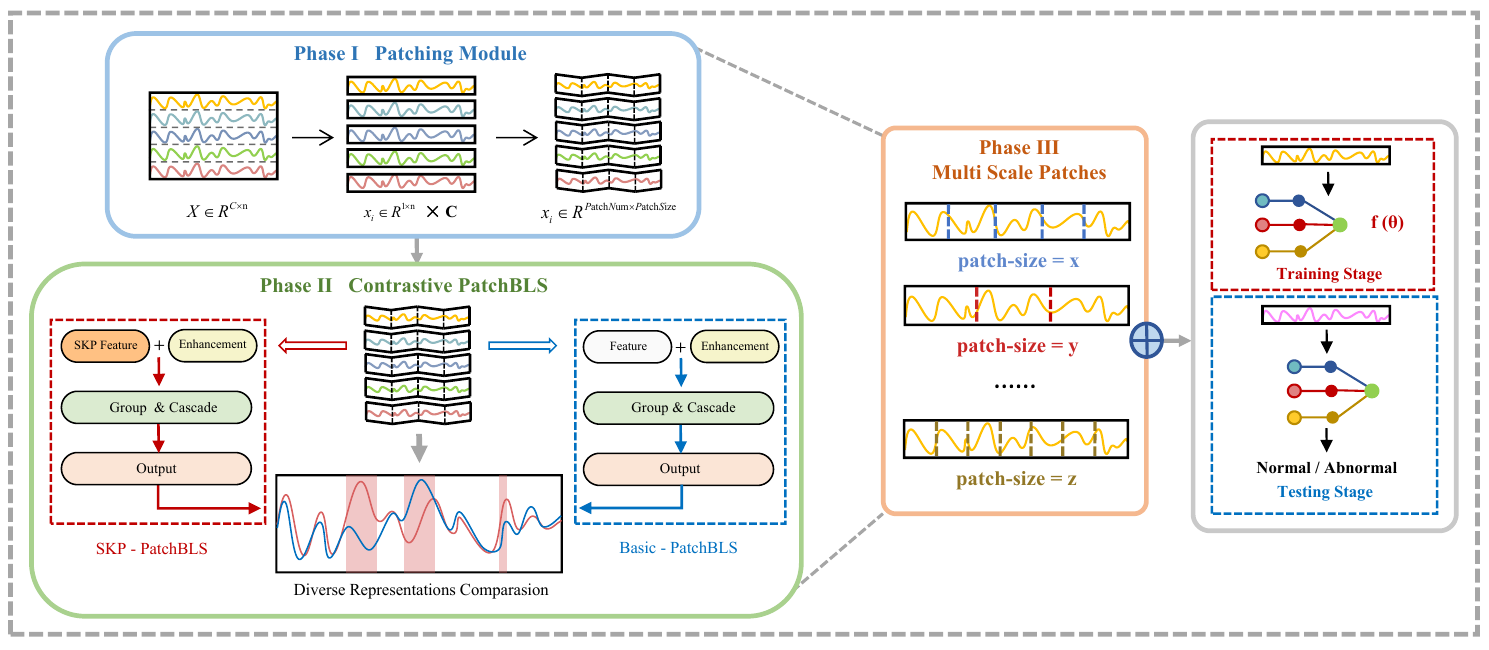}
\caption{Overview of CPatchBLS. First, the Patching Module segments multivariate data into univariate time series. Second, SKP-PatchBLS alongside Basic-PatchBLS are developed for dual-branch analysis. This method compares different representations of the same time series data. Third, CPatchBLS with varying patch sizes effectively utilizes temporal information and achieves ensemble learning at the model level.}
\label{1}
\end{figure*}

\section{Proposed Approach}\label{sec:prop_app}
\subsection{Problem Statement}
In the TSAD task, a time series ${X} = [x_1, x_2, \ldots, x_n]$ is given ${X} \in {\mathbb{R}}^{C \times n}$, with $C$ denoting the number of channels and $n$ denoting the time series length. 
The model with parameters $\theta$ is trained on ${X}_{{\text{train}}}$ and anticipated to represent normal patterns.
The model on the testing dataset ${X}_{{\text{test}}}$ predicts the associated labels ${y} = [y_{1}, y_{2}, \ldots, y_{n}]$, where ${y} \in {\mathbb{R}}^{n}$. An abnormality at a time point $i$ is indicated by a value of $y_i = 1$.

\subsection{Framework of CPatchBLS}
The fundamental concept behind our approach is utilizing a more streamlined BLS architecture with accelerated training and inference speeds to address the limitations of deep structures in the TSAD. The approach aims to overcome the time-consuming nature and intricate model structure associated with deep learning approaches. Through the integration of time series patching and contrastive learning, the proposed CPatchBLS significantly bolstered its representational capacity for better TSAD performance. 

Initially, CPatchBLS incorporates concepts from cutting-edge research and introduces the Patching module into BLS tailored for TSAD, thereby enhancing the representation capabilities of BLS for time series data. Subsequently, following the processing of the original time series data by the patching module, it undergoes a dual-branch contrastive learning framework comprising SKP-PatchBLS and Basic-PatchBLS. This framework, by examining different feature representations from diverse perspectives, facilitates the efficient detection of anomalies. Finally, by capitalizing on the time-efficient attributes of BLS and leveraging Multi-Scale Patches, CPatchBLS harnesses semantic information from multi-scale time series, thereby boosting the overall resilience and performance of the model. 
A detailed exposition of CPatchBLS is provided sequentially in this section.

\subsection{PatchBLS}
\begin{figure}[ht]
\centering
\includegraphics[width=0.9\linewidth]{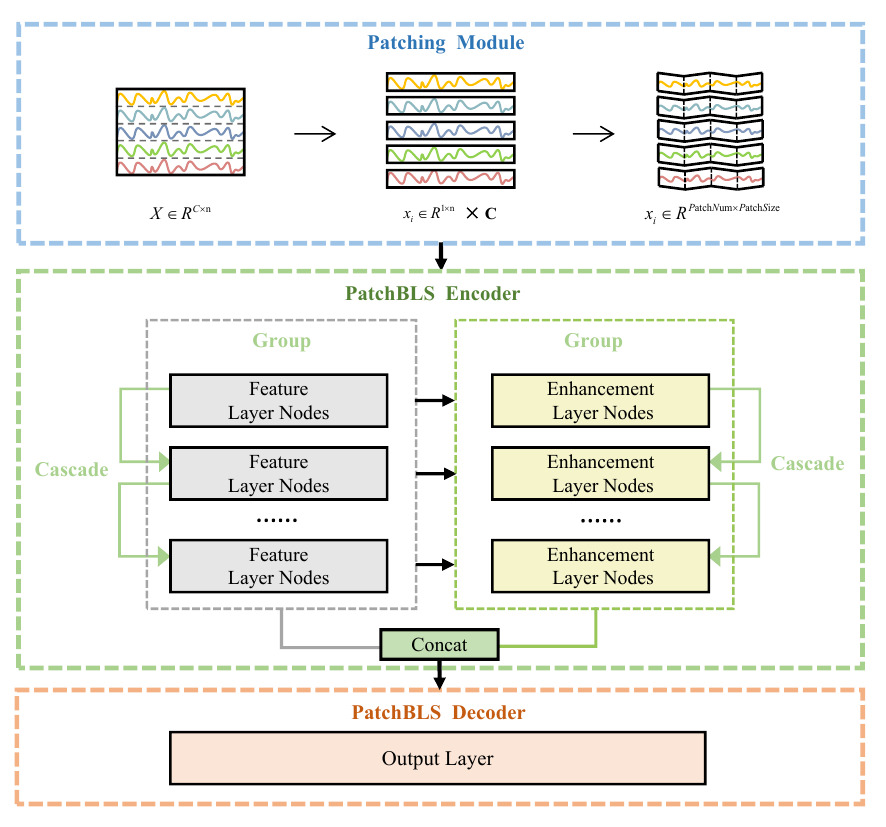}
\caption{The overview of PatchBLS.}
\label{fig:overview_patchbls}
\end{figure}

The temporal patterns inherent in time series data are often highly intricate, making it difficult for simple model architectures to effectively capture and handle these complexities. This challenge has long posed limitations on traditional classical methods, resulting in performance bottlenecks. 
To overcome this challenge and make BLS perceive temporal information, we integrate the patching operation into the BLS framework, resulting in the development of a new paradigm called PatchBLS.
In preprocessing the original time series data for model input, PatchBLS undergoes segmentation into equidistant and non-overlapping time series segments, with each segment referred to as a \textbf{patch}, as shown in Fig. \ref{fig:overview_patchbls}. 
By treating these time series patches as separate entities, the temporal representation capacity is improved due to the fact that local semantic information is preserved inside the time series.
As the basis for PatchBLS to learn sequential semantic information, patches convey information that is not merely a simple individual timestamp. 
Intuitively, PatchBLS is more advantageous than simply examining timestamps alone because it can identify trends and other varied temporal patterns by concentrating on these patches.

The multivariate time series data ${X} \in {\mathbb{R}}^{C\times n}$ first be divided into \(C\) univariate time series ${x}^\prime \in {\mathbb{R}}^{1 \times n}$ in order to comply with the channel independence as described in PatchTST \cite{Patch_PatchTST2022}.
Subsequently, uniform patching is applied to each individual time series sequence, reshaping $x^\prime$ to ${x} \in {\mathbb{R}}^{N_{\text{patch}} \times S_{\text{patch}}}$, where ${N_{\text{patch}}}$ signifies the number of patches and ${S_{\text{patch}}}$ signifies the length of patch.

The subsequent PatchBLS performs the time series patches through PatchBLS's encoder, which is composed of arrays of feature nodes and enhancement nodes. Let \(Z_i\) denote the hidden state of $i$-th feature node, which is represented by the equation: 
\begin{equation}Z_i = \phi \left( {x} W_{i} + \beta_{i} \right), \quad \forall  i \in \{1, 2, \cdots, m\}.
\label{eq:1}
\end{equation}

where $\phi(\cdot)$ represents the activation function, such as ReLU or Sigmoid, $x$ represents the input data, $W_{i} \in {\mathbb{R}}^{S_{\text{patch}} \times D_{\text{ft}}}$ represents the weight matrix, and $\beta_{i} \in {\mathbb{R}}^{S_{\text{patch}} \times D_{\text{ft}}}$ represents the bias matrix, $D_{\text{ft}}$ signifying the dimension of feature node.

The aforementioned weight and bias matrices are all randomly generated and then adapted to the data using the sparse autoencoder method (SAE) \cite{SAE} to optimize these weights. 
The weight matrices are then forced to be orthogonal by the applied Schmid orthogonalization, which lessens the redundancy of various feature node groups.

In order to improve the representation and mining capabilities for time series data, PatchBLS incorporates cascade operations in feature nodes.
The feature nodes are concatenated sequentially, with the output of the previous layer becoming the input of the next layer. This process continues through \(k\) cascade feature layers to delve deeper into the semantic aspects of temporal data. The \(k\)-th cascade feature layer is represented as: 
\begin{equation}
\resizebox{\linewidth}{!}{%
$Z^k_i = 
\begin{cases} 
    \phi \left( x W^1_{i} + \beta^1_{i} \right), & \forall i \in \{1, 2, \ldots, m\}, \, k = 1, \\ 
    \phi \left( Z^{k-1}_i W^k_{i} + \beta^k_{i} \right), & \forall i \in \{1, 2, \ldots, m\}, \, \forall k \in \{2, \ldots, q\}.
\end{cases}%
$}\label{eq:2}
\end{equation}

Additionally, empirical validation suggests that the hidden states of \(m\) feature node groups and the \(k\) cascade layers of feature nodes are commonly concatenated to obtain a more complex representation: 
\begin{equation}
Z = [Z^1_1,\cdots,Z^q_1,\cdots,Z^1_m,\cdots,Z^q_m].
\label{eq:3}\end{equation}

The enhancement nodes are derived from the feature nodes by undergoing the additional linear mappings. The following is the equation when \(n\) groups of enhancement nodes are created:
\begin{equation}H_j = \xi \left( Z {W}_{j} + \beta_{j} \right), \quad \forall j \in \{1, 2, \cdots, n\}.\label{eq:4}\end{equation}

The meanings of the different parameters in the equation are akin to those in Equation (\ref{eq:1}), where \(\xi (\cdot)\) also represents the activation function, \(Z \in \mathbb{R} ^ {N_{\text{patch}} \times (D_{\text{ft}} \times C_{\text{ft}} \times G_{\text{ft}})} \) represents the input feature layer data, \( W_{j} \in \mathbb{R} ^ {(D_{\text{ft}} \times C_{\text{ft}} \times G_{\text{ft}}) \times D_{\text{enh}}} \) represents the weight matrix, and \(\beta_{j} \in \mathbb{R} ^ {(D_{\text{ft}} \times C_{\text{ft}} \times G_{\text{ft}}) \times D_{\text{enh}}} \) represents the bias matrix, with $G_{\text{ft}}$ signifying the group numbers of the feature layers, $C_{\text{ft}}$ signifying the numbers of the cascade feature layers, and $D_{\text{enh}}$ signifying the dimension of the enhancement layer. 

Similarly, we also introduce cascading within the enhancement layer. The \(v\)-th enhancement layer is represented as: \begin{equation}
\resizebox{\linewidth}{!}{%
$H^v_j = 
\begin{cases} 
    \xi \left( Z W^1_{j} + \beta^1_{j} \right), & \forall j \in \{1, 2, \ldots, n\}, \, v = 1, \\ 
    \xi \left( H^{v-1}_j W^v_{j} + \beta^v_{j} \right), & \forall j \in \{1, 2, \ldots, n\}, \, \forall v \in \{2, \ldots, p\}. 
\end{cases}%
$}\label{eq:5}
\end{equation}

There are a total of \(n\) enhancement node groups and \(v\) cascade enhancement layers. Similarly, these enhancement nodes are concatenated, obtaining ${H}^{n} \in {\mathbb{R}}^{N_{\text{patch}} \times (D_{\text{enh}} \times C_{\text{enh}} \times G_{\text{enh}})}$, where $C_{\text{enh}}$ signifying the numbers of the cascade enhancement layer, $G_{\text{enh}}$ signifying the groups of the enhancement layer, and the equation is as follows:
\begin{equation}
H = [H^1_1,\cdots,H^p_1,\cdots,H^1_n,\cdots,H^p_n].
\label{eq:6}
\end{equation}

The next step in PatchBLS is to concatenate all the obtained feature nodes and the enhancement nodes to form the latent features $A \in {\mathbb{R}}^{N_{\text{patch}} \times (T_{\text{ft}} + T_{\text{enh}})}$, where $T_{\text{ft}} = D_{\text{ft}} \times C_{\text{ft}} \times G_{\text{ft}}$ signifying the dimension of the feature node groups, $T_{\text{enh}} = D_{\text{enh}} \times C_{\text{enh}} \times G_{\text{enh}}$ signifying the dimension of the enhancement node groups, which can be expressed by the following:
\begin{equation}{A=[Z, H].}\label{eq:7}
\end{equation}

Deep neural networks rely on backpropagation to iteratively update weights, and they are sensitive to learning rate when network depth increases. PatchBLS, however, uses the pseudo-inverse to quickly compute hidden-to-output weight, ensuring efficient training without issues like the local optima problem \cite{Research_BLS}. 
At this point, the output of this problem can be modeled as: 
\begin{equation}
{Y} = {AW_o}.
\label{eq:8}
\end{equation}
where ${W_o}$ represents the weight matrix of the output block. Since the generation process of the PatchBLS encoder's parameters and the latent feature $A$ are fixed and known and ${Y} \in {\mathbb{R}}^{N_{\text{patch}} \times S_{\text{patch}}}$ are also known as the final target of output, the only parameters that need to be updated in PatchBLS are the weight matrix ${W_o} \in {\mathbb{R}}^{(T_{\text{ft}} + T_{\text{enh}}) \times S_{\text{patch}}}$. Therefore, the objective function of PatchBLS can be expressed as follows:
\begin{equation}
L=\|Y-\hat{Y}\|_{2}^{2}+\lambda\|W_{o}\|_{2}^{2}.
\label{eq:9}
\end{equation}

where ${Y}$ represents the actual values, $\hat{{Y}}$ represents the reconstructed values, the weight matrix that requires training is ${W}_o$, and $\lambda$ is the regularization coefficient preventing overfitting. The pseudo-inverse approach is used to update ${W}_o$ in the manner described below in order to minimize $L$:
\begin{equation}
W_{o}=(A^{\top}A+\lambda I)^{-1}A^{\top}Y.
\label{eq:10}\end{equation}

in which ${I}$ is the identity matrix, and the meanings of other parameters are the same in Equations \eqref{eq:8} and \eqref{eq:9}.

In order to identify anomalies, PatchBLS uses the anomaly score \(\operatorname{Score}=\|Y-\hat{Y}\|^{2}\) during inference. The model generates anomaly predictions for a specific detection threshold, denoted as \(\delta\). This occurs when the score at the time point \(i\) exceeds \(\delta\).

As shown in Fig. \ref{fig:overview_patchbls}, the encoding layer with feature nodes and enhancement nodes in the PatchBLS is utilized as the PatchBLS encoder, while the corresponding output layer is employed as the PatchBLS decoder to generate the representation of time series data through a reconstruction-based approach. However, the reconstruction-based method's limited sensitivity to normal patterns hampers its ability to differentiate between normal and anomalous data \cite{approach_feng2024sensitivehue}. 
Inspired by \cite{CL_DCdetector2023}, we propose Dual-PatchBLS to address these difficulties. We utilize contrastive learning with Dual-PatchBLS to learn a more discriminative perspective, which successfully distinguishes between normal and anomalous data and contributes to better TSAD.

\subsection{Dual-PatchBLS}

Although PatchBLS demonstrates the ability to construct consistent representations of normal patterns, to thoroughly capture temporal features, it is essential to introduce additional view representations and the differences in anomalous data under a contrastive view.
Therefore, we propose a dual-branch contrastive framework, \textbf{Dual-PatchBLS}.

To construct the Dual-PatchBLS, we incorporate random feature approximation transformations within the feature layer of the PatchBLS framework, thereby constructing an abnormal perspective, referred to as the Simple Kernel Perturbation (SKP)-PatchBLS branch. Through an approximation of the Gaussian kernel, more uncertainty is injected into the original PatchBLS while also enhancing the ability of this branch to process nonlinear time series data, capturing more complex temporal data patterns and differing significantly from the original representation. The specific equation of SKP is as follows: 
\begin{equation}\begin{aligned}
\operatorname{SKP}(Z)& =\frac{1}{\sqrt{d_k}}[\sqrt{2}\cos(\omega_{1}^{\top}Z_{}+b_{1}), \\&\ldots,\sqrt{2}\cos(\omega_{d_k}^{\top}Z_{}+b_{d_k})].
\end{aligned}\label{eq:11}
\end{equation}

where $d_k$ denotes the dimension of the random feature space, the variable $Z$ represents time series data following processing by the feature layer, $w_i \sim \mathcal{N}(0, \sigma^2)$ represents the Gaussian random vector, and $b_i \sim U[0, 2\pi]$ represents the random variable.

The preliminary time series anomaly detection result is obtained by comparing the representation results of the above two types of BLS branches with different perspectives and calculating the difference score between \( \text{PatchBLS}_{\text{SKP}}(\cdot) \) and \(\text{PatchBLS}_{\text{Basic}}(\cdot) \) as follows: 
\begin{equation}
\resizebox{0.85\linewidth}{!}{%
$ \text{Score}_{\text{diff}} = \mathbf{D}(\text{PatchBLS}_{\text{SKP}}(X), \text{PatchBLS}_{\text{Basic}}(X)). $
}\label{eq:12}
\end{equation}

Where \( \text{Score}_{\text{diff}} \) represents the difference score between two branches, \( X \) represents the identical time series data, and \( \mathbf{D}(\cdot) \) represents the function calculating the feature difference utilizing \( \text{KL} \) divergence to measure the distinction in our implementation: 
\begin{equation}
\begin{aligned}
\text{Score}_{\text{diff}} &= \frac{1}{2}\text{KL}(\text{PatchBLS}_{\text{SKP}}(X), \text{PatchBLS}_{\text{Basic}}(X)) \\
&\quad + \frac{1}{2}\text{KL}(\text{PatchBLS}_{\text{Basic}}(X), \text{PatchBLS}_{\text{SKP}}(X)).
\end{aligned}%
\label{eq:13}
\end{equation}

\begin{figure*}[ht]
\centering
\begin{minipage}{\textwidth}
\begin{algorithm}[H]
\caption{Basic-PatchBLS's and SKP-PatchBLS's Pytorch-like pseudo-codes.}
\begin{algorithmic}
\label{alg:1}
\STATE \textsc{\textbf{Input:}} The time series data $X \in \mathbb{R}^{C \times n}$; the identical parameters of Basic-PatchBLS and SKP-PatchBLS as follows: the length of the patch \(S_{patch}\), the $D_{\text{ft}}$, $C_{\text{ft}}$, $G_{\text{ft}}$, $D_{\text{enh}}$, 
$C_{\text{enh}}$, $G_{\text{enh}}$, 
the shrink coefficient $s$ of the enhancement layer, and the output regularization $r$ of the output layer; the kernel dimension $d_k$ and the sigma $\sigma$ of SKP-PatchBLS;
\STATE \textsc{\textbf{Output:}} The reconstructed representation $O$ obtained from Basic-PatchBLS or SKP-PatchBLS;
\STATE \hspace{0.5cm} \textbf{for} $x^\prime \in \mathbb{R}^{1 \times n}$ in $X \in \mathbb{R}^{C \times n}$ \textbf{do}
\STATE \hspace{1.0cm} Reshape $x^\prime \in \mathbb{R}^{1 \times n}$ into $x^\prime \in {\mathbb{R}}^{N_{\text{patch}} \times S_{\text{patch}}}$;
\STATE \hspace{1.0cm} Randomly generate the parameters of $G_{\text{ft}}$ groups and $C_{\text{ft}}$ cascade layers of feature nodes;
\STATE \hspace{1.5cm}\textbf{for} $i = 1$ \textbf{to} $m$ \textbf{and} $k = 1$ \textbf{to} $q$ \textbf{do}
\STATE \hspace{2.0cm} Calculate the output of the feature nodes using \eqref{eq:2};
\STATE \hspace{2.0cm} \textbf{if} it's the SKP-PatchBLS branch
\STATE \hspace{2.5cm} Process the output of the feature nodes using \eqref{eq:11};
\STATE \hspace{2.0cm} \textbf{end if}
\STATE \hspace{1.5cm} \textbf{end for}
\STATE \hspace{1.0cm} Connect the feature nodes as \eqref{eq:3};
\STATE \hspace{1.0cm} Randomly generate the parameters of $G_{\text{enh}}$ groups and $C_{\text{enh}}$ cascade layers of enhancement nodes;
\STATE \hspace{1.5cm}\textbf{for} $j = 1$ \textbf{to} $n$ \textbf{and} $v = 1$ \textbf{to} $p$ \textbf{do}
\STATE \hspace{2.0cm} Calculate the output of the enhancement nodes using \eqref{eq:5};
\STATE \hspace{1.5cm} \textbf{end for}
\STATE \hspace{1.0cm} Connect the enhancement nodes as \eqref{eq:6};
\STATE \hspace{1.0cm} Connect the feature nodes and the enhancement nodes to obtain \(A\), as in \eqref{eq:7};
\STATE \hspace{1.0cm} Feed \(A\) to the decoder and calculate the weight \(W\) by \eqref{eq:10};
\STATE \hspace{1.0cm} Obtain the reconstructed representation $O$ by \eqref{eq:8};
\STATE \hspace{0.5cm} \textbf{return} \(O\);
\end{algorithmic}
\label{alg:alg1}
\end{algorithm}
\end{minipage}
\end{figure*}

\subsection{Multi Scale Patches}
The performance of CPatchBLS may vary with different patch size settings when utilizing fixed-length patches for extracting local semantic information from time series data, potentially resulting in reduced model robustness. This assertion is substantiated through subsequent ablation experiments in this study. Moreover, within the TSAD scenario, the manifestation patterns of time series anomalies exhibit notable distinctions, encompassing both point-wise anomalies and more intricate pattern-wise anomalies. Failure to extract and comprehend time series patches' information across multiple scales leads to incomplete exploration of the time series data. This phenomenon is investigated in the experiment.

Given the aforementioned issues, a model-level \textbf{Multi Scale Patches} (MSP) integration method is proposed. This approach leverages temporal information more effectively while preserving the computational efficiency and fast processing capabilities of PatchBLS. By incorporating temporal features from multi-scale patches, the proposed method showcases resilience against the fragility and instability found in individual models. It also covers a broader range of temporal pattern features, effectively tackling significant challenges in time series analysis without adding substantial overhead to the model's inherent strengths. 
The corresponding equation is: \begin{equation}
\resizebox{\linewidth}{!}{%
$ \text{Score}_{\text{Anom}} = \text{Avg}(\text{Score}_{\text{diff}_{1}}, \ldots, \text{Score}_{\text{diff}_{i}}), \quad \forall i \in \{1, 2, \ldots, n\}. $
}\label{eq:14}
\end{equation}
Here $\text{Score}_{\text{Anom}}$ represents the anomaly score of time series and $\operatorname{Avg}(\cdot)$ represents the aggregation method for the difference scores from models with different $n$ multi-scale patch sizes. The variation in $i$ signifies a specific index corresponding to different patch sizes, resulting in different Dual-PatchBLS to achieve varying levels of $\text{Score}_{\text{diff}_{i}}$.

\begin{algorithm}[H]
\caption{CPatchBLS's Pytorch-like pseudo-code.}
\begin{algorithmic}
\STATE \textsc{\textbf{Input:}} The time series data $X \in \mathbb{R}^{C \times n}$; the list of patches $L_{patches}$ and the various parameters related to Basic-PatchBLS and SKP-PatchBLS in Algorithm \eqref{alg:1};
\STATE \textsc{\textbf{Output:}} The anomaly score $\text{Score}_{\text{Anom}}$ obtained from CPatchBLS;
\STATE \hspace{0.5cm} \textbf{for} \(Patch\) in $L_{patches}$ \textbf{do}
\STATE \hspace{1.0cm} Obtain $\text{PatchBLS}_{\text{Basic}}(X)$ and $\text{PatchBLS}_{\text{SKP}}(X)$ using Algorithm \eqref{alg:1};
\STATE \hspace{1.0cm} Obtain the difference score $\text{Score}_{\text{diff}}$ by \eqref{eq:13};
\STATE \hspace{0.5cm} Obtain the anomaly score $\text{Score}_{\text{Anom}}$ by \eqref{eq:14};
\STATE \hspace{0.5cm} \textbf{return} $\text{Score}_{\text{Anom}}$;
\end{algorithmic}
\label{alg:alg1}
\end{algorithm}

\section{Experimental Setting \& Results}
\subsection{Datasets}
Our models and other baselines are evaluated utilizing five widely used real-world datasets: (1) MSL (Mars Science Laboratory dataset) \cite{MSL_dataset}, (2) SMAP (Soil Moisture Active Passive dataset) \cite{SMAP_dataset}, (3) SWaT (Secure Water Treatment testbed) \cite{SWaT_dataset}, (4) WADI (WAter DIstribution testbed) \cite{WADI_dataset}, (5) PSM (Pooled Server Metrics) \cite{PSM_dataset}.

The characteristics of these datasets are presented in Table \ref{tab:datasets_summary}. Entities indicate that the dataset comprises multiple subsets. Continuing from the previous setup \cite{CL_DCdetector2023}, we combine them for model training. Train and Test indicate the number of data time points in the final training and test sets.
\begin{table}[h]
    \centering
    \caption{Summary of Datasets}
    \begin{tabular}{lrrrrr}
        \toprule
        Dataset & Entities & Dims & Train  & Test  & Anomaly Rate (\%) \\
        \midrule
        MSL     & 27      & 55   & 58317    & 73729   & 10.48             \\
        SMAP    & 55      & 55   & 140825   & 44035   & 12.83             \\
        SWaT    & 1       & 51   & 495000   & 449919  & 12.14             \\
        WADI    & 1       & 123  & 1209601  & 172801  & 5.71              \\
        PSM     & 1       & 25   & 132481   & 87841   & 27.76             \\
        \bottomrule
    \end{tabular}
    \label{tab:datasets_summary}
\end{table}

\subsection{Metrics}
In our experiments, we utilize AUC-ROC, AUC-PR, and PA-F1 for a comprehensive evaluation, aligning with recent state-of-the-art works like \cite{ICDE_chen2021daemon}, \cite{approach_2022_tranad}, \cite{ICDE_chen2024learning}, and \cite{ICDE_fang2024temporal}. AUC-ROC and AUC-PR assess the performance of an anomaly detection model across all thresholds. The ROC curve illustrates a binary classifier's diagnostic ability by plotting TPR against FPR. In contrast, the precision-recall curve is beneficial for imbalanced datasets, graphing precision against recall at different thresholds. PA-F1 is suitable for point anomalies and shows robust performance, even with imprecise labels. These multifaceted metrics ensure the accuracy and robustness of the evaluation.
\subsection{Baselines}

In our experiments, we introduced twelve classic and advanced TSAD methods and compared our proposed CPatchBLS in terms of anomaly detection capability, training and testing efficiency, and robustness. From the perspective of the core motivation, we compared classical machine learning methods\footnote{\url{https://github.com/yzhao062/pyod}} such as LOF \cite{ml_lof2000}, IForest \cite{ml_IsolationForest2008}, and ECOD \cite{approach_2022_ecod}, and the machine learning enhanced method Deep SVDD \cite{approach_2018_deep}, reconstruction-based methods like USAD \cite{approach_2020_usad} and FITS \cite{approach_2024_FITS}, Transformer-based methods such as Anomaly Transformer (AnomTrans) \footnote{\url{https://github.com/thuml/Anomaly-Transformer}} \cite{CL_AnomalyTransformer2021} and TranAD \cite{approach_2022_tranad}, 2D-variation model like TimesNet \cite{approach_2022_timesnet}, LLM-based method like OFA \cite{approach_2023_OFA}, contrastive learning-based models like COUTA \cite{approach_2024_COUTA}, and prediction-based method such as LSTMAD \cite{approach_2015_LSTMAD}.
The implementations of Deep SVDD, TranAD, TimesNet and COUTA are publicly available in DeepOD Library\footnote{\url{https://github.com/xuhongzuo/DeepOD }}, and we use the implementations of USAD, FITS, OFA and LSTMAD in TSB-AD repository\footnote{\url{https://github.com/TheDatumOrg/TSB-AD}} \cite{liu2024elephant}.

\subsection{Performance Comparisons}

\begin{table*}[t]
\centering
\caption{Performance comparison results on MSL, SMAP, SWaT, WADI, and PSM datasets.}
\setlength{\tabcolsep}{1mm}
\resizebox{0.95\linewidth}{!}{
\begin{tabular}
{l|ccc|ccc|ccc}
\toprule
\textbf{Dataset}  & \multicolumn{3}{c|}{\textbf{MSL}} & \multicolumn{3}{c|}{\textbf{SMAP}} & \multicolumn{3}{c}{\textbf{SWaT}} \\ \midrule
\textbf{Methods} & \textbf{ROC-AUC} & \textbf{ROC-PR} & \textbf{PA-F1} & \textbf{ROC-AUC} & \textbf{ROC-PR} & \textbf{PA-F1} & \textbf{ROC-AUC} & \textbf{ROC-PR} & \textbf{PA-F1} \\ \midrule
LOF [2000]            & 95.02 & 73.09 & 74.41    & 95.92 & 87.69 & 86.08    & 92.08 & 84.24 & 82.32 \\
IForest [2008]        & 98.03 & 86.16 & 85.76    & 91.88 & 74.74 & 70.18    & 94.04 & 88.62 & 85.54 \\
ECOD [2022]           & 98.20 & 91.42 & 87.55    & 86.71 & 69.36 & 70.30    & 95.72 & 90.02 & 84.94 \\
LSTMAD [2015]         & 99.64 & 96.20 & 93.26    & 94.73 & 82.77 & 78.16    & 88.83 & 80.56 & 82.10 \\
Deep SVDD  [2018]     & 99.14 & 90.44 & 89.63    & 89.69 & 76.30 & 75.75    & 88.29 & 83.49 & 86.58 \\
USAD [2020]           & 90.50 & 57.90 & 64.80    & 85.33 & 68.44 & 68.44    & 79.04 & 38.83 & 53.17 \\
TranAD [2022]         & 94.23 & 74.04 & 73.74    & 93.55 & 77.14 & 69.77    & 77.63 & 38.44 & 53.26 \\
AnomTrans [2022]      & 97.90 & 93.86 & 93.97    & 99.48 & 97.07 & 96.48    & 99.95 & 99.64 & 96.89 \\
TimesNet [2023]       & 99.41 & 91.69 & 91.56    & 88.56 & 77.21 & 73.02    & 99.19 & 96.97 & 93.26 \\
OFA [2023]            & 98.22 & 90.55 & 86.66    & 96.61 & 87.43 & 81.09    & 98.88 & 96.88 & 94.43 \\
COUTA [2024]          & 99.24 & 92.16 & 92.08    & 94.87 & 84.85 & 76.61    & 50.15 & 12.17 & 21.70 \\
FITS [2024]           & 98.26 & 86.52 & 85.65    & 98.48 & 91.53 & 84.02    & 99.68 & 98.38 & 94.45 \\
\midrule
CPatchBLS         & \textbf{99.68} & \textbf{97.54} & \textbf{95.78} & \textbf{99.86} & \textbf{99.23} & \textbf{97.35} & \textbf{99.90} & \textbf{99.28} & \textbf{97.05} 
\\ \midrule
\textbf{Dataset}  & \multicolumn{3}{c|}{\textbf{WADI}} & \multicolumn{3}{c|}{\textbf{PSM}} & \multicolumn{3}{c}{\textbf{Average}} \\ \midrule
\textbf{Methods} & \textbf{ROC-AUC} & \textbf{ROC-PR} & \textbf{PA-F1} & \textbf{ROC-AUC} & \textbf{ROC-PR} & \textbf{PA-F1} & \textbf{ROC-AUC} & \textbf{ROC-PR} & \textbf{PA-F1} \\ \midrule
LOF [2000]           & 95.82 & 78.85 & 73.92    & 93.34 & 84.79 & 78.94    & 94.44 & 81.73 & 79.13 \\
IForest [2008]       & 93.26 & 62.65 & 66.32    & 98.54 & 96.98 & 91.60    & 95.15 & 81.83 & 79.88 \\
ECOD [2022]          & 95.64 & 68.74 & 72.13    & 98.24 & 96.52 & 91.07    & 94.90 & 83.21 & 81.20 \\
LSTMAD [2015]        & 99.49 & 92.45 & 87.72    & 98.39 & 96.96 & 93.05    & 96.22 & 89.79 & 86.86 \\
Deep SVDD [2018]     & 97.66 & 78.91 & 76.57    & 99.52 & 99.11 & 96.72    & 94.86 & 85.65 & 85.05 \\
USAD [2020]          & 95.91 & 73.53 & 70.87    & 95.79 & 92.27 & 86.60    & 89.31 & 66.19 & 68.78 \\
TranAD [2022]        & 98.39 & 75.51 & 79.93    & 96.70 & 93.19 & 84.50    & 92.10 & 71.66 & 72.24 \\
AnomTrans [2022]     & 89.88 & 78.25 & 84.75    & 99.05 & 98.65 & 98.16    & 97.25 & 93.50 & 94.05 \\
TimesNet [2023]      & 98.18 & 74.14 & 82.28    & 99.79 & 99.48 & 97.29    & 97.03 & 87.90 & 87.48 \\
OFA [2023]           & 98.28 & 77.83 & 78.87    & 99.57 & 99.03 & 96.54    & 98.31 & 90.34 & 87.52 \\
COUTA [2024]         & 99.40 & 91.20 & 83.99    & 95.74 & 91.73 & 85.05    & 87.88 & 74.42 & 71.88 \\
FITS [2024]          & 99.08 & 81.98 & 82.93    & 99.75 & 99.41 & 97.12    & 99.05 & 91.56 & 88.83 \\
\midrule
CPatchBLS          & \textbf{99.81} & \textbf{95.57} & \textbf{95.73} & \textbf{99.81} & \textbf{99.63} & \textbf{98.47} & \textbf{99.81} & \textbf{98.25}  & \textbf{96.87}\\ \midrule
\end{tabular}
}
\label{tab:results}
\end{table*}

Table \ref{tab:results} presents a performance comparison between the CPatchBLS and multiple baselines using five real-world datasets. CPatchBLS performs superior to other models in most evaluation metrics. Specifically, compared to the best baseline models on each dataset, CPatchBLS exhibits absolute improvements of 0.77\% in ROC-AUC (99.05\% to 99.81\%), 5.08\% in ROC-PR (93.50\% to 98.25\%), and 3.00\% in F1 score (94.05\% to 96.87\%).

CPatchBLS performed particularly outstandingly on datasets like WADI, which is due to the better data quality and the ability of the model to better reflect real-world anomalies, unlike the MSL and SMAP datasets, which are composed of multiple concatenated sub-datasets rather than continuous ones, resulting in CPatchBLS not being able to establish a larger gap compared to other algorithms. 

Simple classical machine learning methods sometimes outperform some deep learning methods on specific datasets or metrics. For instance, in terms of ROC-AUC, ROC-PR, and PA-F1, the performance of IForest is superior to that of USAD on the MSL dataset. This once again addresses the theme of the article, namely that redundant or complex structures may not necessarily be entirely suitable for TSAD. Due to the strong dependence on priors and the fragility of the representational capabilities of machine learning methods, they are unable to further improve the performance, making them still weaker than advanced deep learning methods. This further demonstrates the necessity of our work bridging machine learning and deep learning.

\subsection{Time Cost Studies}\label{sec:time_cost}

To validate the high efficiency of CPatchBLS, we roughly conducted a comprehensive multi-trial experiment to compare the training and testing time of twelve algorithms with our Ours-Parallel Execution (PE) and Ours-Sequatial Execution (SE) models across five real-world datasets. Ours-PE involves training and testing the Dual-PatchBLS sub-models independently and in parallel within CPatchBLS. The time taken by Ours-PE is measured based on the training and testing duration of the longest Dual-PatchBLS sub-model. Similarly, Ours-SE involves sequentially training and testing the Dual-PatchBLS sub-models within CPatchBLS. Each sub-model is trained and tested one after the other, continuing until all sub-models have been processed.

The average training and testing times of Ours-PE and Ours-SE outperform all deep learning methods in the comparative experiments, strongly confirming the superiority of BLS architecture in shorter time consumption compared to deep learning architectures in the TSAD domain. The average training time (33.6s) and testing time (2.2s) of Ours-SE surpasses the average training and testing time of LOF (55.0s/44.1s). The theoretically optimal model Ours-PE demonstrates even better performance, with average training time (12.6s) and testing time (1.1s) approaching the performance time of ECOD (10.4s/9.4s). Overall time consumption is only lower than IForest (1.7s/0.6s), showing that our proposed method has the potential to match or even surpass some traditional machine learning methods in time consumption. 

\begin{figure}[ht]
\centering
\includegraphics[width=0.75\linewidth]{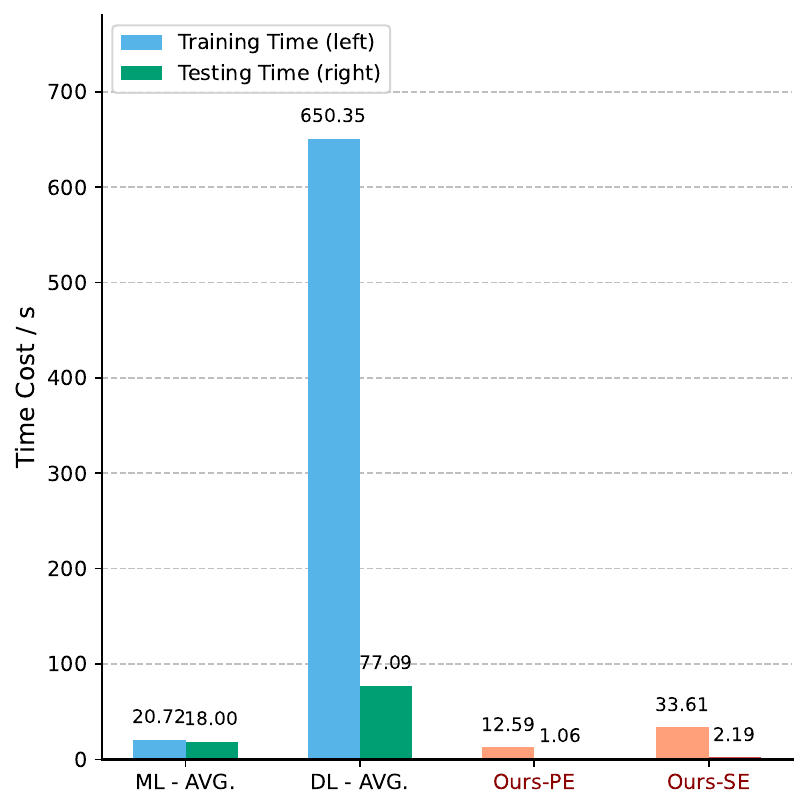}
\caption{Comparison of average time consumption of machine learning methods (ML-AVG.) and deep learning methods (DL-AVG.) with our proposed method reveals significant performance advantages.}
\label{fig:cpatchblsoverview}
\end{figure}

\begin{figure*}[ht]
\centering
\includegraphics[width=1\linewidth]{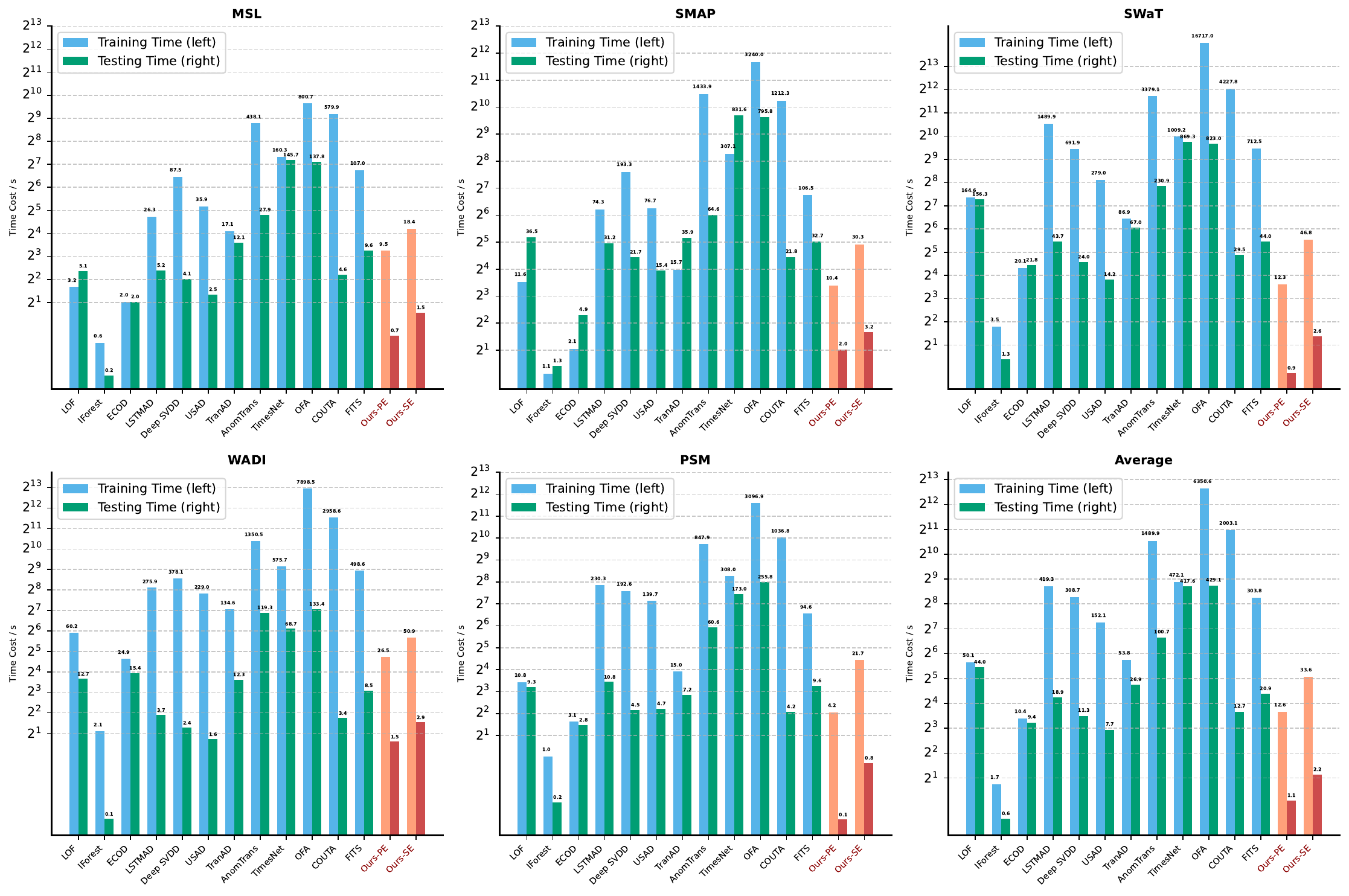}
\caption{Average Training and Testing Time Comparison Results on MSL, SMAP, SWAT, WADI, and PSM Datasets.}
\label{fig:time_cost}
\end{figure*}

In order to more visually demonstrate the advantages of our methods in terms of time consumption, we calculated and compared the average time consumption of various machine learning methods, various deep learning methods, and our method on multiple datasets. To avoid unfair comparison, when calculating the average training and testing time consumption of deep methods, we excluded the OFA model based on GPT2, as its dependence on the characteristics of the LLM has resulted in a significant difference in time consumption compared to other deep learning methods. As shown in Fig. \ref{fig:time_cost}, our two methods, Ours-PE and Ours-SE, both exhibit performance close to the average time consumption of machine learning, far exceeding the average time consumption performance of deep learning methods.

\subsection{Complexity Analysises}
To further demonstrate the time efficiency of our models, we provide a theoretical analysis of the time complexity to further corroborate the superiority. Assuming the input multivariate time series data is $\mathbf{X} \in \mathbb{R}^{C \times n}$, we sequentially analyze the specific time complexities of PatchBLS (SKP-PatchBLS), Dual-PatchBLS, and CPatchBLS:

The primary computation of PatchBLS is the generation of feature nodes and enhancement nodes, as well as the computation of the pseudoinverse. Specifically, each single-channel time series data $\mathbf{x}' \in \mathbb{R}^{1 \times n}$ undergoes transformation by patching into $\mathbf{x}' \in \mathbb{R}^{N_{\text{patch}} \times S_{\text{patch}}}$. Define $n_1$ and $d_1$ as the number and dimension of feature nodes, and $n_2$ and $d_2$ as the number and dimension of enhancement nodes. The computational complexity of generating feature nodes is $T_\text{F} = O(N_{\text{patch}} \cdot S_{\text{patch}} \cdot n_1 \cdot d_1)$. For generating enhancement nodes, it is $T_\text{E} = O(N_{\text{patch}} \cdot n_1 \cdot d_1 \cdot n_2 \cdot d_2)$. Hence, the total computational complexity of node generation can be expressed as:
\begin{equation}
\begin{aligned}
    T_\text{FE} &= T_\text{F} + T_\text{E} \\
       &= O(N_{\text{patch}} \cdot S_{\text{patch}} \cdot n_1 \cdot d_1) + O(N_{\text{patch}} \cdot n_1 \cdot d_1 \cdot n_2 \cdot d_2) \\
       &= O(N_{\text{patch}} \cdot \max\{S_{\text{patch}}, n_2 \cdot d_2\} \cdot n_1 \cdot d_1).
\end{aligned}
\end{equation}

According to \ref{eq:10}, the complexity of the pseudo-inverse calculation is:
\begin{equation}
\begin{aligned}
T_\text{W} &= O(\max\{N_{\text{patch}}, 2d\} \cdot \min\{N_{\text{patch}}, 2d\}^2) \\
&= O(N_{\text{patch}} \cdot (2d)^2) \\
&= O(N_{\text{patch}} \cdot d^2).
\end{aligned}
\end{equation}
where $d = n_1 \cdot d_1 + n_2 \cdot d_2$. Collectively, considering the overall complexity of PatchBLS performing $C$ channels is:
\begin{equation}
\begin{aligned}
T_\text{PatchBLS} &= C \cdot ( T_\text{FE} + T_\text{W}) \\
&= O(C \cdot N_{\text{patch}} \cdot \max\{S_{\text{patch}}, n_2 \cdot d_2\} \cdot n_1 \cdot d_1) \\ 
&\quad+ O(C \cdot N_{\text{patch}} \cdot d^2). \\
\end{aligned}
\end{equation}

The introduction of the random feature approximation operation for minor data perturbation in the SKP-PatchBLS structure, which maps the data post-feature layer to a nonlinear space of $d_k$ dimensions, implies that, based on the available information, its time complexity can be determined as:
\begin{equation}
\begin{aligned}
T_\text{SKP-PatchBLS} &= O(C \cdot N_{\text{patch}} \cdot \max\{S_{\text{patch}}, n_2 \cdot d_2\} \cdot n_1 \cdot d_1 \cdot d_k) \\ 
&\quad + O(C \cdot N_{\text{patch}} \cdot (n_1 \cdot d_1 \cdot d_k + n_2 \cdot d_2)^2). \\
\end{aligned}
\end{equation}

When \(d^2\) is sufficiently large, \(T_{\text{PatchBLS}}\) can be considered as \(O(C \cdot N_{\text{patch}} \cdot d^2)\). Similarly, \(T_{\text{SKP-PatchBLS}}\) can be approximated as  \(O(C \cdot N_{\text{patch}} \cdot (n_1 \cdot d_1 \cdot d_k + n_2 \cdot d_2)^2)\).

Dual-PatchBLS operates by running Basic-PatchBLS and SKP-PatchBLS independently and then combining their results to determine difference scores. Therefore, its time cost is associated with the execution of these two branches as follows:
\begin{equation}
\begin{aligned}
T_{\text{Dual-PatchBLS}} &= T_{\text{PatchBLS}} + T_{\text{SKP-PatchBLS}} \\
&= O(C \cdot N_{\text{patch}} \cdot d^2) \\
&\quad+ O(C \cdot N_{\text{patch}} \cdot (n_1 \cdot d_1 \cdot d_k + n_2 \cdot d_2)^2) \\
&= O(C \cdot N_{\text{patch}} \cdot (n_1 \cdot d_1 \cdot d_k + n_2 \cdot d_2)^2).
\end{aligned}
\end{equation}

As CPatchBLS is an ensemble model comprising multiple Dual-PatchBLS with varying patch sizes,  the ultimate time complexity of CPatchBLS can be approximated as:
\begin{equation}
\begin{aligned}
T_\text{CPatchBLS} ={}& O(C \cdot N_{\text{patch}_1} \cdot (n_1 \cdot d_1 \cdot d_k + n_2 \cdot d_2)^2) \\
                & {} + O(C \cdot N_{\text{patch}_2} \cdot (n_1 \cdot d_1 \cdot d_k + n_2 \cdot d_2)^2) \\
                & {} + \cdots \\
                & {} + O(C \cdot N_{\text{patch}_n} \cdot (n_1 \cdot d_1 \cdot d_k + n_2 \cdot d_2)^2).
\end{aligned}
\end{equation}

Therefore, the computational complexity of Ours-SE mentioned in Sec. \ref{sec:time_cost} is $T_\text{CPatchBLS}$, while the parallel computational complexity of Ours-PE depends on the computational bottleneck. Therefore, its complexity is as follows: 
\begin{equation}
    T_\text{Ours-PE} = \max_i \{ {O(C \cdot N_{\text{patch}_i} \cdot (n_1 \cdot d_1 \cdot d_k + n_2 \cdot d_2)^2)} \}.
\end{equation}

\subsection{Ablation Studies}

\begin{table}[th]
\centering
\caption{Ablation study performances of proposed method on MSL, SMAP, SWaT, WADI, and PSM datasets.}
\resizebox{\linewidth}{!}{
\begin{tabular}
{l|ccc|ccc|ccc}
\toprule
\textbf{Dataset}  & \multicolumn{3}{c|}{\textbf{MSL}} & \multicolumn{3}{c|}{\textbf{SMAP}} & \multicolumn{3}{c}{\textbf{SWaT}} \\ \midrule
\textbf{Methods} & \textbf{ROC-AUC} & \textbf{ROC-PR} & \textbf{PA-F1} & \textbf{ROC-AUC} & \textbf{ROC-PR} & \textbf{PA-F1} & \textbf{ROC-AUC} & \textbf{ROC-PR} & \textbf{PA-F1} \\ \midrule
PatchBLS              & 99.49 & 93.75 & 91.58   & 89.57 & 73.83 & 70.64    & 73.39 & 40.68 & 53.38 \\
SKP-PatchBLS          & 99.54 & 94.54 & 92.71   & 99.62 & 96.46 & 96.35    & 97.28 & 87.08 & 85.98 \\
Dual-PatchBLS  & 99.47 & 94.81 & 92.89   & 99.83 & 98.42 & 96.43    & 99.73 & 98.03 & 94.26 \\
CPatchBLS             & 99.68 & 97.54 & 95.78   & 99.86 & 99.23 & 97.35    & 99.90 & 99.28 & 97.05 \\
\midrule
\textbf{Dataset}  & \multicolumn{3}{c|}{\textbf{WADI}} & \multicolumn{3}{c|}{\textbf{PSM}} & \multicolumn{3}{c}{\textbf{Average}} \\ \midrule
\textbf{Methods} & \textbf{ROC-AUC} & \textbf{ROC-PR} & \textbf{PA-F1} & \textbf{ROC-AUC} & \textbf{ROC-PR} & \textbf{PA-F1} & \textbf{ROC-AUC} & \textbf{ROC-PR} & \textbf{PA-F1} \\ \midrule
PatchBLS              & 96.25 & 71.26 & 65.15   & 96.47 & 94.48 & 89.24   & 91.03 & 74.80 & 74.00 \\
SKP-PatchBLS          & 90.16 & 58.14 & 63.46   & 99.48 & 98.52 & 96.05   & 97.22 & 86.95 & 86.91 \\
Dual-PatchBLS  & 98.26 & 77.73 & 76.03   & 99.68 & 99.31 & 97.02   & 99.40 & 93.66 & 91.33 \\
CPatchBLS             & 99.81 & 95.57 & 95.73   & 99.81 & 99.63 & 98.47   & 99.81 & 98.25 & 96.87 \\
\midrule
\end{tabular}
}
\label{tab:abls}
\end{table}
We conducted ablation studies on the PatchBLS base model, the SKP module (SKP-PatchBLS),  the contrastive learning module (Dual-PatchBLS), and the multi-scale patch model integration module (CPatchBLS) on all datasets, with the specific results shown in Table \ref{tab:abls}. We can draw the following conclusions from the experimental data: 

\begin{enumerate}
    \item The overall performance of SKP-PatchBLS is comprehensively superior to PatchBLS under the average evaluation results on multiple datasets and metrics, proving that time series have complex nonlinear temporal patterns and the original PatchBLS capture capability is still insufficient. The introduction of the nonlinear representation helps to better complete the TSAD task. 
    \item In most cases (except on the WADI dataset), PatchBLS outperforms SKP-PatchBLS. This shows that the introduction of random kernel approximation transformation brings in more uncertain factors but does not fully guarantee the model's robustness, resulting in performance loss. The Dual-PatchBLS compares the above two-branch model. It shows stable performance that surpasses the two models. This demonstrates that contrastive learning enhances the robustness and stability of the overall framework and further improves TSAD performance.
    \item The performance of CPatchBLS completely exceeds that of the above models. This indicates that introducing multi-scale patches has improved the model's ability and its mining of temporal information, while reducing the information loss of single-scale patch.
\end{enumerate}

\begin{figure*}[!t]
\centering
\includegraphics[width=0.95\linewidth]{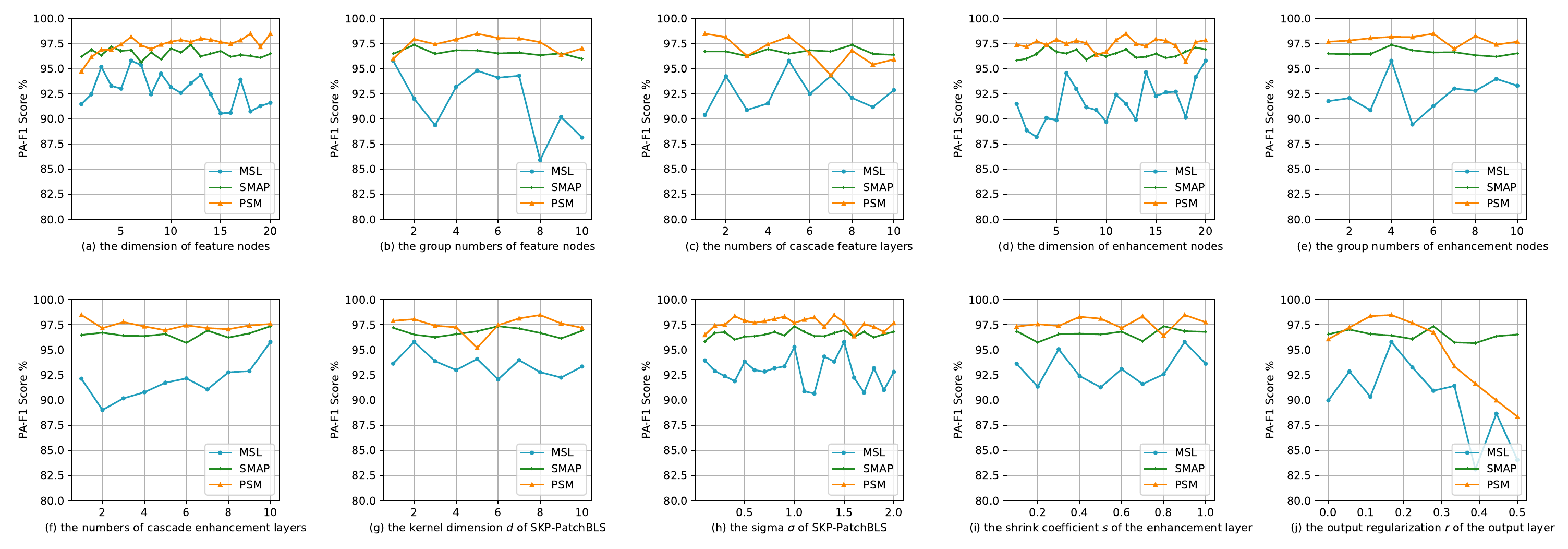}
\caption{Sensitivity analysis of the proposed method on MSL, SMAP, and PSM datasets.}
\label{fig:sens}
\end{figure*}

\begin{figure}[!ht]
\centering
\includegraphics[width=0.7\linewidth]{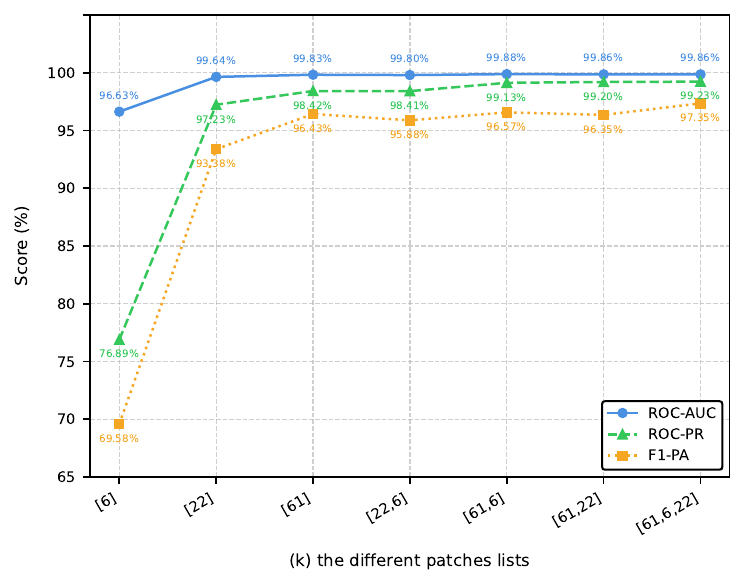}
\caption{Sensitivity analysis w.r.t patch lists of the proposed method on SMAP datasets.}
\label{fig:sen_patch}
\end{figure}

\subsection{Parameters Sensitivity Studies}
To analyze the sensitivity of CPatchBLS to parameter settings, we conduct sensitivity experiments with different parameters on the MSL, SMAP, and PSM datasets.
The results of a series of parameter sensitivity experiments reveal significant findings. Across three datasets, the number of groups in the feature layer, the number of cascade layers in the feature layer, and the number of groups in the enhancement layer all presented a distinct peak within the range of [4, 8]. Meanwhile, the performance improvement observed with a deeper enhancement layer suggests that, although CPatchBLS does not require an excessively deep network structure like traditional deep learning, incorporating groups or cascade mechanisms can significantly enhance its TSAD capabilities and overall performance. These findings collectively validate that introducing groups and cascade mechanisms can optimize the network structure and improve the model's performance. However, an excessive number of redundant nodes and overly deep network layers may lead to a drastic performance decline due to challenges in capturing low-frequency temporal features. Moreover, the ideal shrink coefficient \(s\) typically falls between 0.8 and 1.0, and the output regularization \(r\) is usually within the range [0.1, 0.3], as excessively high regularization coefficients can significantly degrade model performance. 

As shown in Fig. \ref{fig:sen_patch}, a comprehensive sensitivity experiment was conducted on the SMAP dataset for various patch lists. 
To explore the influence of Dual-PatchBlS with different patch sizes on the final CPatchBLS, we choose different patch lists to construct CPatchBLS.
It was found that the single model Dual-PatchBLS is highly sensitive to patch size selection. There are notable performance discrepancies at patch sizes of 6, 22, and 61, highlighting its reliance on patch size selection. In contrast, the ensemble of multiple patches demonstrates enhanced stability, underscoring CPatchBLS's robustness compared to Dual-PatchBLS. 

When two different patch sizes are combined, the performance is better than when only one patch size is used. When three patch sizes of [61, 6, 22] are used simultaneously, consistent improvements are achieved in most metrics.

\section{Conclusion}
We propose CPatchBLS, an innovative perspective that offers the advantages of generalized anomaly detection performance and time efficiency. CPatchBLS utilizes the PatchBLS structure to effectively characterize and mine semantic information within complex temporal data. By constructing SKP-PatchBLS with small kernel perturbations and combining it with the original PatchBLS branch, a comparative paradigm of Dual-PatchBLS is constructed to represent the same temporal data from different perspectives, further improving the temporal anomaly detection performance. Finally, we introduced the Multi Scale Patches mechanism and further improved the overall performance of CPatchBLS through the idea of ensemble learning. The overall performance and various components of CPatchBLS have been effectively validated through extensive experiments. The experimental results show that the proposed CPatchBLS outperforms twelve baselines in anomaly detection performance and time costs on five real-world benchmarks from different scenarios.

\section*{Acknowledgments}
This work was supported in part by the National Natural Science Foundation of China 62476101, 62106224, U21A20478, 62222603, 62076102, in part by the National Key R\&D Program of China 2023YFA1011601, in part by the Key-Area Research and Development Program of Guangdong Province under number 2023B0303030001, and in part by Guangzhou Science and Technology Plan Project 2024A04J3749 and the Fundamental Research Funds for the Central Universities (2024ZYGXZR062).

\bibliography{approach,dataset,history_work,introduction}
\bibliographystyle{IEEEtranN}

\end{document}